\documentclass[letterpaper, 10 pt, conference]{ieeeconf}  %

\IEEEoverridecommandlockouts                              %

\usepackage{balance}
\usepackage{amsmath}
\usepackage{wrapfig,lipsum,booktabs}
\usepackage{amssymb}
\usepackage{mathtools}

\usepackage{graphicx} %
\graphicspath{ {./imgs/} }
\usepackage{hyperref}
\usepackage{pgfplots}
\pgfplotsset{width=10cm,compat=1.9}
\pgfplotsset{every axis/.append style={
                    axis line style={<->}, %
                    xlabel={$x$},          %
                    ylabel={$y$},          %
                    label style={font=\small},
                    tick label style={font=\tiny},
                    title style={font=\tiny},
                    width=0.27\textwidth,
                    height=0.27\textwidth,
                    legend style={font=\small},
                    legend pos=south east,
                    }}
\newcommand{\para}[1]{\vspace{0.2em}\noindent\textbf{#1}}

\usepackage{xcolor}

\title{\LARGE \bf
Probable Object Location (POLo) Score Estimation for \\Efficient Object Goal Navigation 
}

\author{Jiaming Wang$^{1}$ and Harold Soh$^{1,2}$%
\thanks{$^{1}$Jiaming Wang and Harold Soh are with Smart Systems Institute, National University of Singapore.  
        {\tt\small jamie-w@nus.edu.sg}}%
\thanks{$^{2}$Dept of Computer Science, National University of Singapore. 
        {\tt\small harold@comp.nus.edu.sg}}%
}

\begin{document}

\maketitle
\thispagestyle{empty}
\pagestyle{empty}

\begin{abstract}
To advance the field of autonomous robotics, particularly in object search tasks within unexplored environments, we introduce a novel framework centered around the Probable Object Location (POLo) score. Utilizing a 3D object probability map, the POLo score allows the agent to make data-driven decisions for efficient object search. We further enhance the framework's practicality by introducing POLoNet, a neural network trained to approximate the computationally intensive POLo score. Our approach addresses critical limitations of both end-to-end reinforcement learning methods, which suffer from memory decay over long-horizon tasks, and traditional map-based methods that neglect visibility constraints.
Our experiments, involving the first phase of the OVMM 2023 challenge, demonstrate that an agent equipped with POLoNet significantly outperforms a range of baseline methods, including end-to-end RL techniques and prior map-based strategies. To provide a comprehensive evaluation, we introduce new performance metrics that offer insights into the efficiency and effectiveness of various agents in object goal navigation.
\end{abstract}

\section{INTRODUCTION}

Efficiently searching for objects in unexplored environments is a crucial skill for intelligent agents in real-world scenarios. Humans excel at solving this partially observable task by employing two key strategies: 1) leveraging common object associations, such as finding a coffee mug near a coffee machine~\cite{sarch_tidee_2022, narasimhan_seeing_2020}; 2) employing efficient search behaviors like glancing into unexplored rooms while passing by and thoroughly examining promising areas where the target object is likely situated. This study concentrates on the latter aspect: devising a method to equip autonomous robots with efficient search behaviors.

Current state-of-the-art methods for this task can be broadly classified into two categories: map-based methods and non-map-based methods. Non-map-based methods~\cite{maksymets_thda_2021, ramrakhya_habitat-web_2022, ramrakhya_pirlnav_2023} often rely on end-to-end learning paradigms, leveraging implicit memory structures like RNNs or LSTM networks to handle the partial observability of the environments. While effective in simple scenarios where the target objects are conspicuous and the environment is limited in size, these methods encounter difficulties in complex environments. It has been shown that information stored in the memory cell in baseline LSTMs decays exponentially~\cite{mahto2020multi, tallec2018can}, potentially leading to a decline in spatial awareness for the agent. This results in agents re-exploring areas they have already traversed, ultimately causing a failure to locate the target object within the allocated time constraints.

Conversely, map-based methods tackle the challenge by concurrently constructing a spatial representation of the environment as the robot explores. This constructed map serves as an explicit memory for the agent, providing the advantage of enhanced generality and interpretability. However, it also necessitates the development of efficient search strategies. Existing search methodologies in this area often adopt a two-stage heuristic: initial map exploration followed by object-specific search. However, these methods typically overlook critical considerations such as what is visible from a specific map position~\cite{ramakrishnan_poni_2022, chaplot_object_2020} and prioritized exploration of promising areas~\cite{heng_efficient_2015, bircher_receding_2016, selin_efficient_2019, schmid_efficient_2020}, which in turn leads to suboptimal search trajectories.

\begin{figure*}
  \centering
  \includegraphics[width=\textwidth]{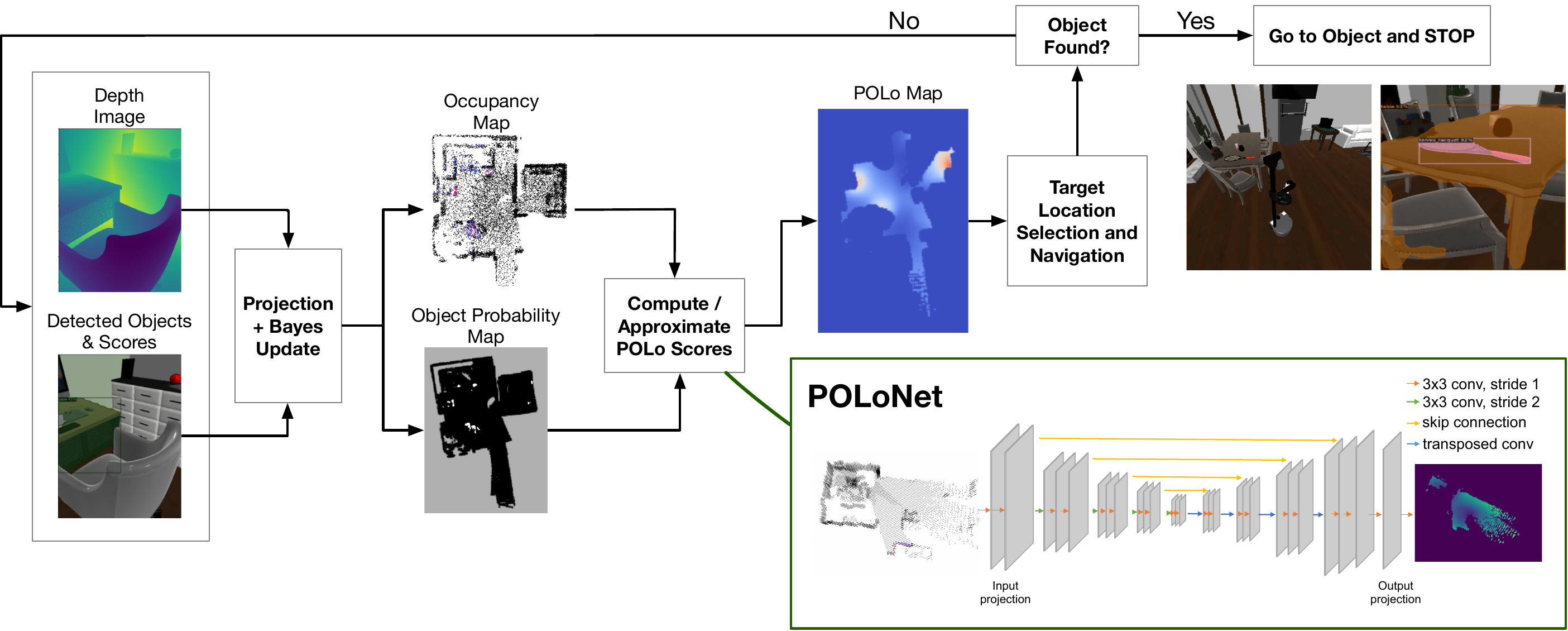}
  \caption{Overview of our POLo Robot Agent for Object Goal Navigation. Here, the robot is tasked to find a tennis racquet on a table. The robot has a depth camera and uses an object detector (DETIC~\cite{zhou_detecting_2022} in our case). It maintains an occupancy map and an object probability map, which are used to compute a Probable Object Location (POLo) score for each robot configuration --- this score reflects the agent's belief of an object being located within the viewable areas from a certain configuration. Since computing the POLo score is expensive (it requires computing which voxels are visible), we  approximate it using POLoNet, a prediction network that takes as input a tensor with dimensions $15\times 192\times 192$, where the height dimension (15) serves as the channel dimension for 2D convolution. This tensor is created through cropping and down-sampling of the current global occupancy map $\mathcal{M}^o$ and the probabilistic map $\mathcal{M}^p$. POLoNet produces an output tensor of size $2\times 192\times 192$, representing a dense POLo score map. The robot selects the configuration with the highest POLo score as the next target location and navigates to the area.}
  \label{fig:overview}
\end{figure*}

In this paper, we introduce a map-based framework to address the above limitations (Fig. \ref{fig:overview}). Central to our framework is the introduction of a new metric called the \emph{Probable Object Location} (POLo) score. The POLo score serves to: (i) quantify the likelihood of an object being located within a 3D environment, thus guiding the agent to focus on high-probability areas while still taking into account visibility constraints; (ii) promote the exploration of as-yet-unmapped regions. Despite its utility, computing the POLo score can be computationally expensive as it requires consideration of visible areas (akin to rendering).

To address this computational bottleneck, we introduce an approximation technique using a trained neural network that we term POLoNet. This neural network is  engineered to estimate the POLo score efficiently, without significantly compromising accuracy. The integration of POLoNet within our framework enables a robot to make real-time decisions in complex environments, thereby mitigating the computational cost  associated with such comprehensive scores.

We validate our framework using the first phase of the OVMM 2023 challenge, and our empirical results are promising. An agent equipped with the POLoNet framework significantly outperforms various baseline approaches; it achieves better performance comapred to an end-to-end Reinforcement Learning (RL) method and previously proposed map-based strategies. Specifically, our agent demonstrates enhanced performance in terms of success rate and efficiency metrics like Success weighted by Path Length (SPL). This evidence substantiates the robustness and effectiveness of our approach, the POLo score and POLoNet. In brief, our main contributions are as follows:
\begin{itemize}
\item We propose a map-based framework for efficiently searching for objects in an unknown environment. To the best of our knowledge, this is the first work to incorporate a ``next-best-view''-like strategy for object goal navigation tasks. Experiments show that our POLoNet framework significantly outperforms other baseline methods.
\item We conduct a comprehensive set of experiments and analyses on the proposed method and multiple baseline methods. We introduce new metrics for evaluating object navigation tasks, providing insights into the behavior of alternative methods for object goal navigation.
\end{itemize}

\section{Task Definition}
\label{section:task}
In the Object Goal Navigation \cite{chaplot_object_2020, wani2020multion} task, the agent's objective is to locate an object specified by its object class name. At each step, the agent receives egocentric RGB and depth images, in addition to localization (GPS) and compass readings. The action space consists of four discrete actions: MOVE FORWARD, TURN LEFT, TURN RIGHT, and STOP. An episode is deemed successful when the agent confirms object discovery by executing the STOP action within a certain threshold distance $d_s$ from the target object. The episode terminates after a fixed number of timesteps.

To evaluate and compare different methods, we adopt the object goal navigation sub-task from the 2023 Habitat Open Vocabulary Mobile Manipulation (OVMM) challenge~\cite{homerobotovmmchallenge2023}. The challenge defines goals using natural language, e.g., ``\emph{move an apple from the table to the counter}''. The OVMM challenge comprises four distinct sub-tasks, where the agent is tasked not only with locating objects as described in text but also with executing subsequent object manipulation tasks, including pick and place actions directed towards specified receptacles. 

For our study, we focus on the object goal navigation phase, representing the initial stage within the OVMM challenge. This phase is difficult; 
unlike the original Habitat object navigation task, the OVMM environment more closely simulates real-world conditions. The OVMM objects are from an open set and are comparatively smaller, such as cups and cellphones, as opposed to the beds and couches. The scenes in the OVMM environment are also larger, compelling the agent to explore its surroundings more efficiently.

\section{Related Works}
\label{section:related works}

\para{Object Goal Search With Implicit Memories.}
There is a growing body of research dedicated to object goal navigation tasks. One research avenue focuses on training agents directly using end-to-end reinforcement learning or imitation learning~\cite{mirowski2018learning,codevilla2018end}. Typically, RNNs or LSTMs serve as implicit memory mechanisms to encode the environment. To enhance agent performance and generalization, Maksymets et al. \cite{maksymets_thda_2021} proposed the use of data augmentation by generating additional training episodes with common household objects, while Ye et al.~\cite{ye_auxiliary_2021} employed auxiliary tasks to facilitate learning. In contrast, Ramrakhya et al.~\cite{ramrakhya_habitat-web_2022} collected a substantial human demonstration dataset and demonstrated that agents trained using behavior cloning on this dataset can acquire efficient searching behaviors. However, end-to-end methods, in general, suffer from low sampling efficiency~\cite{wijmans_dd-ppo_2020} and limited generalizability when applied to real-world scenarios~\cite{ gervet_navigating_2022}. Our experiments reveal that a RL agent tended to revisit already searched areas, potentially due to memory decay issues in its RNN.

\para{Object Goal Search with Explicit Maps.}
Our work utilizes an explicit map to represent the environment, which enables sample efficiency and potentially better generalization via a modular  paradigm~\cite{ramakrishnan_poni_2022, chaplot_object_2020, gupta_cognitive_2019, georgakis_learning_2021}. Prior map-based methods methods use explicit representations as proxies for robot observations and employ either learned or heuristic policies to select high-level goals based on the constructed map. In contrast, our approach incorporates a 3D probabilistic map and a probable object location score, allowing the agent to select object locations in a manner that balances exploration and exploitation. %

\para{Exploration with Next Best View.}
A task that is related to object goal search is autonomous exploration. An effective strategy for this task involves employing a ``next-best-view'' selection strategy aimed at maximizing the coverage of unexplored areas efficiently. This approach optimizes ``information gain'' along the exploration trajectory, as demonstrated in \cite{heng_efficient_2015, bircher_receding_2016, selin_efficient_2019, schmid_efficient_2020}. In these studies, information gain, $\textbf{Gain}(\zeta)$, is defined as 
\begin{equation}
    \textbf{Gain}(\zeta)=\textbf{Visible}(\mathcal{M},\zeta)e^{-\lambda d_\zeta},
\end{equation}
where $\textbf{Visible}(\mathcal{M},\zeta)$ represents the set of visible and unmapped voxels from the configuration $\zeta$ within the global occupancy map $\mathcal{M}$. $d$ is the path cost from the current configuration to $\zeta$, and tuning factor $\lambda$ penalizes high path costs.
This approach facilitates exploration and coverage of the environment, but is not directly applicable to object goal navigation since it does not explicitly inspect  promising regions where the target object is likely to be situated.

\section{Probable Object Location (POLo) for\\Object Goal Navigation}

When searching an object in an unknown environment, humans handle uncertainty by considering potential object locations. For instance, when encountering an unexplored room, we briefly investigate it. If we notice anything indicative of the object's presence, we examine further. Otherwise, we move on to explore other areas. This behavior suggests a representation of the environment that captures \emph{probable object locations} based on previous observations, and a strategy that considers both exploration and exploitation.

With this in mind, we developed an agent equipped with a 3D representation to track probable locations of the object, supplementing the conventional occupancy map used in prior work (Fig. \ref{fig:overview}). To optimize the search using this map, we introduce a measure known as the Probable Object Location (POLo) score. This score quantifies the likelihood of an object's presence within visible voxels, thereby enabling the agent to explore areas of high probability without neglecting visibility constraints. However, calculating the POLo score is computationally expensive. To mitigate this issue, we propose an approximation technique using a trained neural network, termed POLoNet. Subsequent sections elaborate on each of these components, beginning with the probabilistic map, followed by discussions on the POLo score and POLoNet.

\subsection{3D Target Object Probability Map}
\label{section:probabiliy_map}

As stated above, the robot maintains both a 3D object probability map, represented as $\mathcal{M}^p$, and an occupancy map, denoted as $\mathcal{M}^o$. This is in contrast to relying solely on a 2D/3D binary occupancy map as in prior object navigation~\cite{ramakrishnan_poni_2022, chaplot_object_2020, chen_learning_2018}, and map coverage tasks~\cite{heng_efficient_2015, bircher_receding_2016, selin_efficient_2019, schmid_efficient_2020}.  
This 3D probabilistic map represents regions that are likely to contain the target object. It is represented as a tensor with dimensions $\mathcal{M}^p\in \mathcal{R}^{L\times L \times H}$, where $L$ represents the size of the 3D map, and $H$ denotes the height. Each voxel corresponds to a  5cm$^3$ physical space in the environment. 

The map is updated at each time-step; we generate an ego-centric local map $\mathcal{M}_L\in \mathcal{R}^{N\times N \times H}$ using the current depth image and detection results obtained from an open vocabulary detector~\cite{zhou_detecting_2022}. Each entry in the local map corresponds to a detection likelihood $p_L(v_i | o^T) = g(s_\text{goal}, s_\text{recp})$, where $s_\text{goal}$ and $s_\text{recp}$ are the detection scores for the goal object and its receptacle, respectively (as specified by the language instructions). In our work, $g$ is a distance-based weighted combination of the detection scores, but alternative functions can be used. Subsequently, we perform a spatial transformation to convert this local map into the global frame and update the global map using Bayes rule:%
\begin{equation} \label{eq:bayesian}
    \begin{aligned}
        p(v_{i}|o^T,o^{1:T-1})&=  \frac{p(v_{i}|o^{1:T-1})}{1-p(v_{i}|o^{1:T-1})}\times \frac{p(v_{i}|o^T)}{1-p(v_{i}|o^T)} \\
         & \times \frac{1-p(v_{i})}{p(v_{i})} \times (1-p(v_{i}|o^T,o^{1:T-1}))
    \end{aligned}
\end{equation}
where $p(v_{i}|o^T)$ represents the likelihood of an object of interest in voxel $v_i$ given the observation $o$ at timestep $T$, and $p(v_i)$ denotes the prior probability. Note that our map assumes that the probability of the an object in a specific voxel is independent of other voxels (once conditioned upon the observation). Although this may not be true, this independence assumption allows us to efficiently maintain this probabilistic map, as each voxel can be updated independently. 

In preliminary experiments, we attempted to directly use this 3D map directly by selecting the voxel with the highest probability as the target goal location. However, this purely exploitation-oriented approach yielded suboptimal results. The limitations appear to stem from two primary factors: (i) noisy detections affecting the accuracy of the probabilistic map, and (ii) an insufficient exploration strategy that failed to investigate other potentially viable areas. To rectify these shortcomings, we adopted a more balanced methodology, which involves calculating the POLo score, described in the next subsection.

\subsection{Probable Object Location Score (POLo)}
\label{section:POLo}

Using the global 3D object probability map, we calculate the \textbf{Probable Object Location Score (POLo)}. Conceptually, the POLo score gauges the ``utility'' of an \emph{area} surrounding a given robot pose within the map. The score comprises two key elements: (i) an exploration component, which encourages investigation of unexplored regions rich in visible voxels, and (ii) an exploitation component, which favors areas comprising voxels with a high probability of containing the target object. Formally, POLo is defined as,
\begin{align} \label{eq:POLo}
    \text{POLo}(\zeta)=\Big(\sum_{v_i\in \mathcal{V}^\zeta_{<\delta}}f(p(v_i),\zeta) + 
    \beta \sum_{v_i\in \mathcal{V}^\zeta_{\geq\delta}}f(p(v_i),\zeta)\Big)
    e^{- \lambda d_\zeta}
\end{align}
where $\mathcal{V}^\zeta_{<\delta} = \{ v \in \textbf{Visible}(\mathcal{M}^o,\zeta) |  p(v_i) < \delta \}$ and $\mathcal{V}^\zeta_{\ge\delta} = \{ v \in \textbf{Visible}(\mathcal{M}^o,\zeta) |  p(v_i) \ge \delta \}$ and $\textbf{Visible}(\mathcal{M}^o,\zeta)$ represents the set of visible and unmapped voxels from the configuration $\zeta$ within the global occupancy map $\mathcal{M}^o$. The function $f(\cdot,\zeta)$ is used to adjust probabilities based on the voxel's location relative to the agent's configuration. Considering that accuracy in object detection decreases with distance~\cite{hao2022understanding}, we apply a distance-based decay $f(p(v_i),\zeta)=p(v_i)/d(v_i,\zeta)$, where $d(v_i,\zeta)$ represents the distance between voxel $v_i$ and the configuration $\zeta$. 

The POLo score incorporates two summands: the first corresponds to the exploration component, while the second represents the exploitation component. The last exponent multiplier in the equation serves to penalize configurations based on their distance, akin to the \textbf{Gain} measure. Two parameters, $\beta$ and $\delta$, modulate the score's behavior. Specifically, $\beta$ dictates the degree to which high-probability voxels are prioritized during the search. Conversely, $\delta$ represents our confidence level in the voxel-based object probabilities. A higher $\delta$ value fosters a more conservative strategy, as fewer voxels will contribute to the exploitation component in the score calculation.

In our implementation, the agent's next target location was determined by the configuration $\zeta \in \mathcal{R}^2 \in [0,L]\times[0,L]$—representing its position on the map—that yielded the highest POLo score. Although the view angle $\theta$ could potentially be incorporated into the agent's configuration $\zeta$, our preliminary experiments showed that omitting $\theta$ simplifies computation without negatively impacting performance. Therefore, $\textbf{Visible}(\mathcal{M}^o,\zeta)$ includes all voxels visible from $\zeta$, under the assumption that the agent has a panoramic (360°) field of view.

\subsection{POLoNet}
\label{section:POLo_net}

Computing the POLo score is  similar to 3D voxel rendering, with probabilities associated with each voxel as features. For these calculations, we utilized the PyTorch3D library~\cite{ravi2020pytorch3d}. However, calculating POLo scores for all conceivable configurations within the map proved to be excessively time-consuming.

To circumvent this bottleneck, we trained a neural network, termed POLoNet, to predict POLo scores (Fig. \ref{fig:overview}).  
The network takes as input a cropped global voxel map, denoted as $\mathcal{M}^{crop}\in \mathcal{R}^{N'\times N' \times H}$, derived from the global occupancy map $\mathcal{M}^o$ and the probabilistic map $\mathcal{M}^p$. In this tensor, each voxel has value -1 if unobserved or $p(\mathbf{V}_i)$ otherwise. 

Our network architecture design is based on achieving a large receptive field covering the agent's maximum depth range of 10m. This enables POLo score prediction by considering the surrounding 3D structures. We employ a U-Net-like network structure with skip connections, which enables the incorporation of both coarse and fine-grained features. Instead of employing expensive 3D convolution, we use 2D convolutions by treating the height dimension as a channel dimension. The network outputs a dense POLo map $\mathcal{M}^{\text{POLo}}\in \mathcal{R}^{2\times N'\times N'}$, with the two layers in the output tensor corresponding to the two terms in (\ref{eq:POLo}).

As a preliminary evaluation, we trained POLoNet on ground truth labels calculated using Eqn. (\ref{eq:POLo}). Overall, we generated 19,726 data pairs $((\mathcal{M}^o, \mathcal{M}^p),\mathcal{M}^{POLo})$, with each label $\mathcal{M}^{POLo}$ containing POLo scores sparsely rendered at 150 randomly chosen locations due to computational budget limitations. To optimize learning from this constrained dataset, we applied data augmentation techniques such as random translation and rotation during training.

POLoNet was tested with 1,464 novel data pairs. We compared POLoNet against two variants: (i) one without skip connections (w/o SC), and (ii) one without data augmentation (w/o DA). Results in Table \ref{tab:network eval} reveal that POLoNet achieves a low test Mean Absolute Error (MAE). Both skip connections and data augmentation contributed to error reduction. Remarkably, the average computation time for POLoNet stood at only 0.26 $\pm$ 0.06 seconds—a 76-fold speedup compared to the conventional POLo score computation, which required 20 $\pm$ 3.23 seconds\footnote{Computed for a uniform sample of 150 locations within navigable, mapped areas}.

\begin{table}
    \centering
    \caption{POLoNet Error Scores with Ablations}
    \label{tab:network eval}
    \begin{tabular}{lc}
    \toprule
        Method &  Test MAE $\downarrow$ \\
        \midrule
         POLoNet & 0.0117 $\pm$ 0.0468\\
         w/o SC & 0.0258 $\pm$ 0.1270 \\
         w/o DA &  0.0190 $\pm$ 0.1257 \\
         \bottomrule
    \end{tabular}
\end{table}

\section{Experiments and Results}
\label{section:results}

In this section, we report on experiments on object goal navigation using the Habitat simulator \cite{szot_habitat_2021}. Our primary goal was to validate if the POLo score and its estimation using POLoNet led to performance gains. We evaluated POLoNet and other baselines using 200 episodes generated from 12 novel scenes in the test dataset. An episode was successful if the target object was found within 1250 simulation steps, and considered unsuccessful otherwise. 

\subsection{Experimental Setup}
\para{Model Training.} In our experimental evaluation, we trained the POLo prediction network (POLoNet) using 38 scenes from the training dataset using the techniques described in the previous section. 

\para{Baselines and Variants.} We benchmarked the performance of POLoNet against four baseline agents and two variants of POLoNet, each implementing distinct navigation and search strategies:
\begin{itemize}
\item \textbf{RL:} This agent, trained end-to-end with DD-PPO~\cite{wijmans_dd-ppo_2020}, utilizes current depth images and segmentation masks of detected target objects to predict the next discrete action.
\item \textbf{FBE:} Employing a frontier-based exploration strategy~\cite{yamauchi1997frontier}, this agent maintains and updates a 2D semantic map at each step and advances toward the goal object upon identification.
\item \textbf{Max Coverage:} The next high-level goal for this agent is selected based solely on the highest Gain score, calculated from the 3D global occupancy map, in a manner similar to prior work~\cite{bircher_receding_2016}.
\item \textbf{Max Object Probability:} This agent also selects the next high-level goal based solely on the highest probability voxel in the 3D object probability map.
\end{itemize}
and the following variants:
\begin{itemize}
\item \textbf{POLo-2D:} This variant calculates the next high-level goal by considering 2D occupancy and 2D probabilistic maps, obtained by projecting the original 3D maps onto the XY plane.
\item \textbf{POLo-R:} This agent computes the POLo score using PyTorch3D, but restricts computations to a uniform sample of 150 locations within mapped and navigable areas.
\end{itemize}
For RL and FBE agents, we used the  implementation and pre-trained weights from \cite{homerobot}. We also experimented POLoNet agents with different $\beta$ values.

\para{Metrics.} We employed the following metrics to  assess agent performance:
\begin{itemize}
\item \textbf{Success Rate:} This metric represents the ratio of successful episodes to the total number of episodes evaluated.
\item \textbf{SPL} (Success weighted by Path Length): This metric measures the efficiency of the agent in reaching the target object, as outlined in~\cite{anderson2018evaluation}.
\item \textbf{E/D Ratio} (Explored Area divided by Total Traveled Distance): This metric evaluates the agent's proficiency in exploration and map coverage.
\item \textbf{CP/D Ratio} (Closely Checked Promising Area divided by Total Traveled Distance): Promising regions are defined as detected receptacles potentially containing the goal object. This metric quantifies the agent's efficiency in exploitation by assessing the thoroughness in inspecting promising areas.
\end{itemize}
These latter two metrics furnish additional insights into agent performance, complementing the standard metrics conventionally employed in object goal navigation tasks.

\begin{table}[t]
  \caption{Evaluation Metrics on the Habitat Simulator}
  \label{tab:result}
  \centering
  \begin{tabular}{lcccc}
    \toprule
    Method       & Success (\%) $\uparrow$ & SPL $\uparrow$ & E/D $\uparrow$ & CP/D $\uparrow$ \\
    \midrule
    RL & 12.50 & 0.060 & 1.91 & 0.014 \\
    FBE  & 10.99 & 0.049 & 0.98 & 0.010 \\
    Max Coverage  & 12.95 & 0.051 & 3.49 & 0.021 \\
    Max Object Prob. & 10.94 & 0.054 & 3.80 & 0.040 \\
    \hline
    POLo-2D & 13.20 & 0.050 & \textbf{4.48} & 0.033\\
    POLo-R ($\beta=5$) & 17.71 & \ 0.066 & 3.17 & 0.040 \\
    POLoNet ($\beta=1$) & 16.23 & 0.055 & 3.30 & 0.037 \\
    POLoNet ($\beta=3$) & 17.71 & 0.083 & 3.35 & 0.040 \\
    POLoNet ($\beta=5$) &\textbf{ 20.31} & \textbf{0.083} & 3.52 & \textbf{0.044} \\
    \hline
    \bottomrule
  \end{tabular}
  
\end{table}

\begin{figure}
  \centering
  \includegraphics[width=0.47\textwidth]{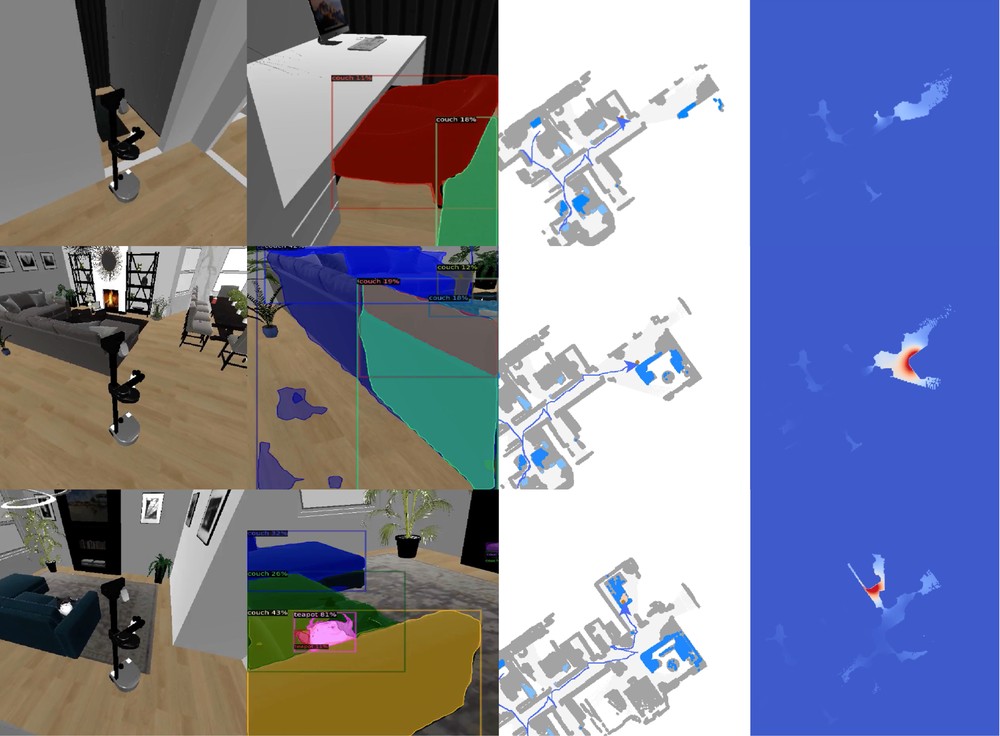}
  \caption{The exploration trajectory (top to bottom) of finding a teapot on a couch. From left, the first image displays the third person view of the agent, the second image is the agent's RGB observation with overlaid object detection results. The third image shows the constructed map, along with the agent's location (blue pointer) and its travelled paths (blue curves). The last image is the predict values from the POLoNet. High values are in red. We observe that the agent is guided by the POLo scores and checks multiple rooms by briefly peeking into it (top). When it detects a couch, the POLo score leads the agent to closely check the couch by predicting high values around the couch (middle). The agent continues to explore the map, and successfully finds the teapot (bottom).}
  \label{fig:trajectory}
\end{figure}

\subsection{Results and Discussion}
\label{section:results discussion}

\para{Comparisons against Baselines.} The results from our experiment are summarized in Table \ref{tab:result}. The proposed POLoNet with $\beta=5$ agent outperforms all the baselines by a significant margin in terms of success rate. Moreover, it demonstrates higher scores in terms of the SPL and CP/D ratios, indicating more efficient navigation and exploitation of the promising regions. Qualitatively, we observed that the POLoNet guides the agent towards check promising areas and exploring  unmapped territory. Fig. \ref{fig:trajectory} visualizes an example trajectory using POLoNet, showing RGB observations, detected objects, projected 2D semantic maps, and predicted POLo maps at different time points in an episode. We can observe that the agent displays efficient search behaviors such as peeking into rooms and closely checking potential receptacles.

\begin{figure}
    \centering  
    \begin{tikzpicture}
        \begin{axis}[
            name=axis1,
            xlabel=Number of Steps,
            ylabel=Total Explored Area,
        ]
        \addplot [blue] table[x=step_num, y=RL_E, col sep=comma] {results.dat};
        \addplot [red] table[x=step_num, y=FBE_E, col sep=comma] {results.dat};
        \addplot [green] table[x=step_num, y=UR_E, col sep=comma] {results.dat};
        \legend{RL,FBE,POLo}
        \end{axis}
        \begin{axis}[
            name=axis2,
            at={(axis1.outer north east)},anchor=outer north west,
            xlabel=Number of Steps,
            ylabel=Total Checked Area,
        ]
        \addplot [blue] table[x=step_num, y=RL_C, col sep=comma] {results.dat};
        \addplot [red] table[x=step_num, y=FBE_C, col sep=comma] {results.dat};
        \addplot [green] table[x=step_num, y=UR_C, col sep=comma] {results.dat};
        \legend{RL,FBE,POLo}
        \end{axis}
    \end{tikzpicture}
    \caption{Left: Accumulated explored (mapped) area at each step, averaged for 200 evaluated episodes. Right: Accumulated closely checked promising area at each step, averaged for 200 evaluated episodes.}
    \label{fig:exp_graph}
\end{figure}
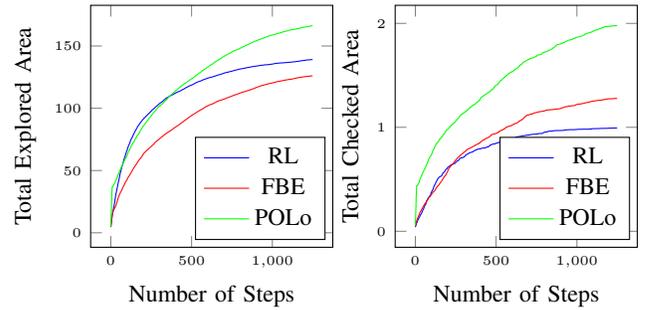

\begin{figure}
  \centering
  \includegraphics[scale=0.25]{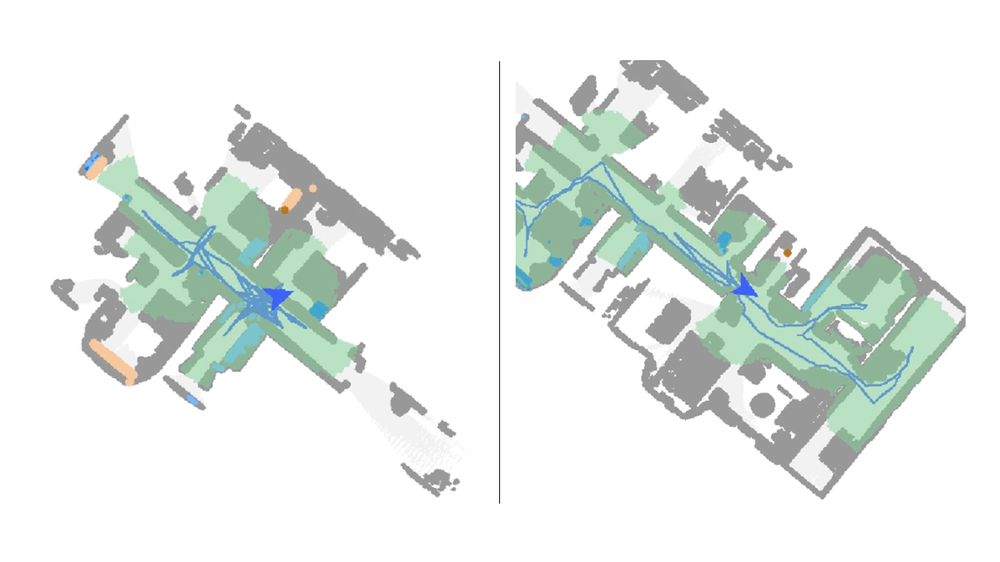}
  \caption{Exploration trajectories represented by blue curves for the RL agent (left) and the POLoNet agent (right), for the same episode. The RL agent explores a smaller area within an equivalent number of steps due to its tendency to repeatedly visit the same room.}
  \label{fig:rl_vs_ur}
\end{figure}

The behavior of the RL agent highlights the challenges associated with long-term planning and exploration in object goal navigation. While the RL agent demonstrates efficient low-level control and good initial exploration, its performance drops in longer episodes. The limitations in its implicit memory system result in repetitive, unproductive exploration patterns. This finding is illustrated in Fig. \ref{fig:exp_graph}, where the exploration rate plateaus after around 800 steps. In contrast, the POLoNet agent maintains a consistent level of efficient exploration throughout the episode, as demonstrated qualitatively in Fig. \ref{fig:rl_vs_ur}. This suggests that the POLoNet provides a more effective framework for exploration and exploitation over long time horizons.

The limitations of the Frontier-Based Exploration (FBE) strategy stem from its lack of contextual awareness about what to observe in the environment. While FBE excels at identifying unexplored areas by locating frontiers, it lacks the sophistication to differentiate between areas with differing levels of importance or information content. This often results in suboptimal behavior, such example, choosing inefficient intermediate goals on the frontier. This highlights the importance of using a more structured and information-rich criterion like POLo for guiding agent behavior, as opposed to relying on more simplistic strategies.

With regards to methods that maximize coverage or maximize object probability, these are strategies that emphasize solely exploration or exploitation, respectively. Neither do particularly well compared to POLoNet, which suggests a middle-ground approach that balances the two criteria is important.

\para{POLoNet Variants.} The 2D POLo agent has  relatively high E/D and CP/D ratios, indicating the advantages of employing POLo scores even in a 2D context. However, its lower success rate compared to the 3D POLoNet variants highlights the limitations of 2D maps in POLo calculations; a 2D-based approach cannot differentiate between a table and a wall. The tuning of the $\beta$ parameter affects the CP/D ratio and success rates; a higher $\beta$ encourages the agent to scrutinize promising areas before venturing into unexplored territories. This is particularly beneficial in expansive environments where transitioning between rooms incurs a higher cost.

\para{Error Analysis.} The success rates of all methods in our experiment are notably lower compared to the success rates reported in the 2002 Habitat Object Goal Navigation Challenge  (HON)~\cite{noauthor_habitat_nodate-1}. This discrepancy arises from the greater difficulty of the OVMM environment. All the agents encounter failures stemming from faulty perception due to small target objects. Notably, the success rate of the POLoNet agent increases to {56.7\%} when employing ground truth perception. In addition, we observed that the presence of numerous narrow corridors and doors within the OVMM scenes frequently leads to collisions and failures in the low-level controller employed by modular approaches. We aim to address these issues as part of our future work. %

\section{Conclusions}
In this study, we introduce a framework enabling autonomous agents to efficiently search for objects in unknown environments. Our proposed method maintains a continuously updated 3D probabilistic map and quantifies Probable Object Location Score (POLo) based on this map to guide the search process. The POLo score provides the agent with a data-driven basis for decision-making, allowing it to focus on high-probability zones without overlooking visibility constraints. Recognizing the computational burden of calculating the POLo score, we introduced POLoNet, a neural network designed to approximate this score efficiently.

Our empirical evaluation, undertaken within the context of the OVMM 2023 challenge, demonstrates that an agent equipped with POLoNet outpaces a variety of baseline methodologies. Our work also contributes to the research community by introducing new performance metrics that allow for a more nuanced understanding of an agent's efficiency and effectiveness in object goal navigation tasks. Moving forward, we are working to transfer these successes to real-world robotic systems, accounting for sensor noise, occlusions, and other real-world challenges.

\section*{ACKNOWLEDGMENT}
This research is supported by the National Research Foundation, Singapore under its Medium Sized Center for Advanced Robotics Technology Innovation.

\balance
\bibliographystyle{IEEEtran}
\bibliography{zot_lib}

\end{document}